# Abstract

In this work, the possibilities for segmentation of cells from their background and each other in digital image were tested, combined and improoved. Lot of images with young, adult and mixture cells were able to prove the quality of described algorithms. Proper segmentation is one of the main task of image analysis and steps order differ from work to work, depending on input images. Reply for biologicaly given question was looking for in this work, including filtration, details emphasizing, segmentation and sphericity computing. Order of algorithms and way to searching for them was also described. Some questions and ideas for further work were mentioned in the conclusion part.

# Acknowledgement

I would like to thank to my supervisors Benny Thörnberg and Dalibor Štys for problem discussions and work encouragement. I would also like to thank to Vítězslav Březina and Štěpánka Kučerová for replying my questions in biological view of the task background. And at last I have to thank to all people and staff of Mid Sweden University in Sundsvall for oportunity to spend some time there, and learn new things.

# 1    Introduction

## 1.1    Task Description

The goal of this work is to analyze the set of images of live HeLa cells, growing for experiments focused on cytotoxity in vitro [1] in Laboratory of tissue cultures, Academic and University Center of Nové Hrady (Institut of Systems Biology and Ecology of Academy of Sciences of the Czech Republic and Institut of Physical Biology of University of South Bohemia) [2].

The cells were observed by inverted microscope OLYMPUS IX51, using phase contrast. Every 2 minutes was taken one image by camera OLYMPUS C-7070.

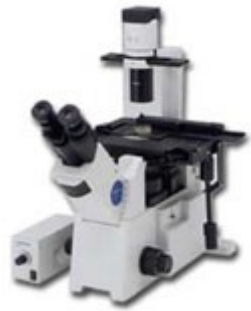

Figure 1.: *Microscope OLYMPUS IX51, source: http://olympuseurope.com*

The background on all images is the bottom of the NUNC, plastic cultivation ware.

Monitoring the cells growth – number of cells in time and their shape specification – are the knowledges wanted to be reached from the set of  images. Because of amount of snapshots (3937 jpeg files), it is time expensive to analyze them by human. Then the algorithms of image processing for automatic analysis were tested and implemented in MATLAB environment [3].

The task could be separated into this main steps, that demand to be solved:

i)    object to background segmentation, simply find where is NUNC and where are HeLa cells

ii)   independent objects location in the image, means move from „know where the cells can be"
      to „where the cells are" and where are the boundaries between them

iii)  computation the output parameters – quantity and sphericity of the found cells

Final solution will be used in praxis in Laboratory of tissue cultures in Nové Hrady for further experiments.

## 1.2    HeLa Cell Line

On 4[th] of October 1951 in Johns Hopkins Hospital in Baltimore died Henrietta Lacks, a black woman in age 31. The reason of her death was cervical cancer. Without her knowledge, the sample of tumor cells were taken by the researches. This cell line survived to the nowadays and is still used in laboratories as a model for human cells in thousands of biological experiments, contributing to the understanding of disease processes.

HeLa cells were propagated into an immortal human cell line by George Otto Gey, scientist from Tissue Culture Laboratory at  Johns Hopkins Hospital. The word "HeLa" was devised by him using the first two letters of Mrs. Lacks' first and last names to keep her real name in a secret.

Lacks' cancer cells have evolved into a self-replicating, single-cell life-form and to HeLa cells were given the new species name: *Helacyton gartleri*. The cells are the genetic chimera of human papillomavirus HPV18 and human cervical cells. [4]

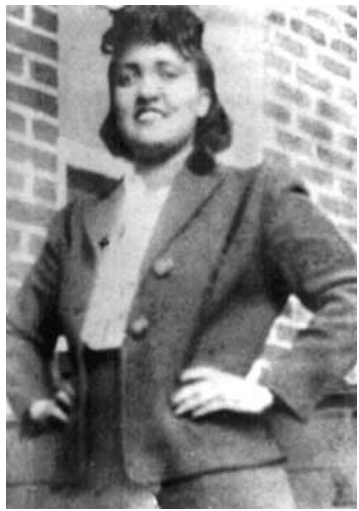

Figure 2.: *Henrietta Lacks, source: http://wikipedia.org/*

## 1.3    Phase Contrast Microscope

The method of phase contrast allow to observe soft, colourless, transparent objects, especially living cells. The phase differences of light beams are passing through the cell and converted  into differences in amplitude. That made them visible for human eye.

The principle of phase contrast for microscopy was first time proposed by Dutch scientist Frederik Zernike in 1932. His idea was awarded with the Nobel price in Physics in 1953.

Between specimen and observer is situated the phase-plate consist of thin annulus, changing the phase by angle $\pi/2$ or $-\pi/2$ for negative or positive phase contrast. For positive phase contrast ($-\pi/2$ phase-plate) appear the thicker parts darker.

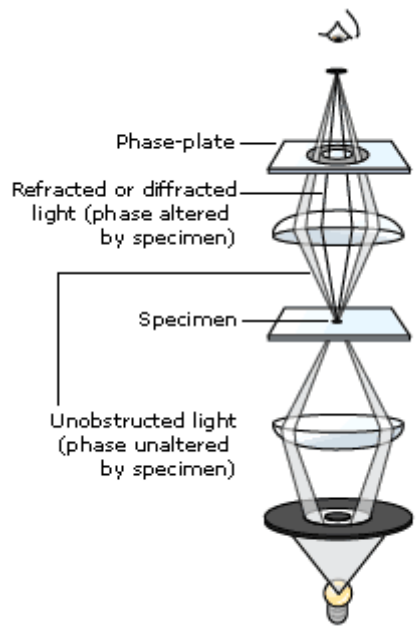

Figure 3.: *Scheme of phase contrast microscope, source: http://nobelprice.org/*

Unfortunately, there are also some important disadvantages. When the specimen was strongly refracted, a halo effect occurs. That means, very shiny boundary overlapped the real object boundaries. The next limitation is disappearing of absorbing saturated objects for the observer.

# 2    Method

Image Analysis is the way how extract hopefully useful informations from the images in automatic or semi-automatic methods and their algorithms on computer processing. Plenty of them are well described in literature [5, 6] or Image Analysis software guide [3, 7].

Sometimes for solving the tasks is enough to use the most common known and third-party preprogrammed algorithms. Then the problem is simply only how to find the proper sequence of this algorithms and their parameters. Some tasks are more difficult and need to discovery new mathematicaly based method. But the main number of real tasks, given by people not involved in Image Analysis, are combination of those two points of view – use already described methods and, where their failed, develop some level of improvement.

In this part, the basic methods, used in whole work, will be shortly mentioned with presumtion of poor previous Image Analysis knowledge.

## 2.1    Colour and grayscale representations

Most common colour representation using by machine for vision or display is RGB (Red, Green, Blue) colour space, where each pixel of image is represented by triplet (r,g,b). Value of colour channel (triplet element) is equal to intensity of its colour. The value for each channel is usually situated in interval <0,255> or <0,1> [5].

| | | | |
|---|---|---|---|
| 0 | 0 | 0 | 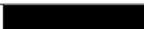 |
| 255 | 0 | 0 | 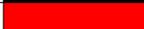 |
| 0 | 255 | 0 | 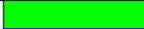 |
| 0 | 0 | 255 | 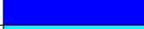 |
| 0 | 255 | 255 | 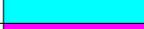 |
| 255 | 0 | 255 | 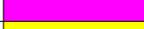 |
| 255 | 255 | 0 | 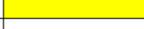 |
| 255 | 255 | 255 | 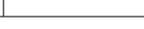 |

Table 1.: *RGB colours*

Grayscale representation is an image in 256 shades of gray <0,255>. There are three ways how to create grayscale image from RGB image. First one is accorded to the relative sensitivity of human eye for primary colours:

$$Y1 = 0.3*r + 0.59*g + 0.11*b$$

In the second one, just combination of intensity of all three channels is done with the same weight coeficient:

$$Y2 = \frac{1}{3}*r + \frac{1}{3}*g + \frac{1}{3}*b$$

The last one is weight channels relative to each other [7]:

$$Y3 = \frac{r^2 + g^2 + b^2}{r + g + b}$$

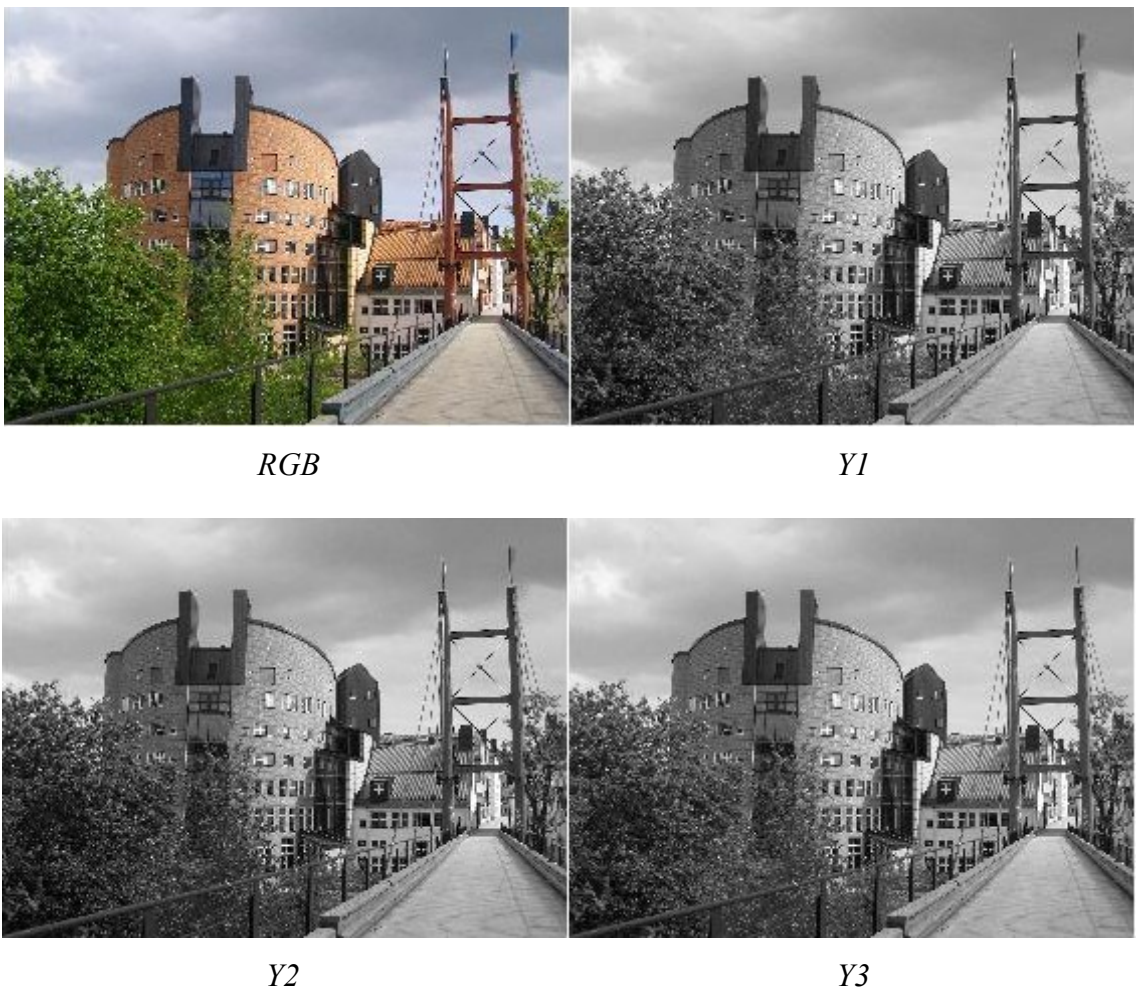

RGB                                        Y1

Y2                                         Y3

Figure 4.: *Mid Sweden University in Sundsvall – RGB and grayscales, source: author*

During the grayscale transformation are the  informations about colour lost and can not be restored back from grayscale image.

## 2.2    Intensity Histogram

The Histogram function $H(p)$ is an intensity function, shows count of pixel $f(i,j)$ with the intenzity equal $p$ independently on the position $(i,j)$.

$$H(p) = \sum_{i,j} h(i,j,p)$$
$$h(i,j,p) = 1 \; if \; f(i,j) = p$$
$$= 0 \; if \; f(i,j) \neq p$$

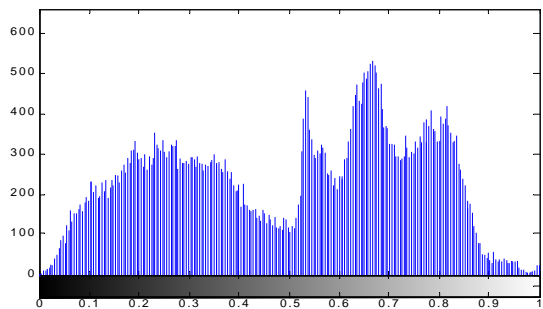

Figure 5: *Histogram function for Y1*

## 2.3    Thresholding

This method is time cheap and that is why it is often in use. It is searching for local minima or maxima in intensity histogram for separating image into the objects related to the real objects. It takes from the image parts that corresponding to the threshold parameter(s). Increasing of threshold parameters may (but have not to) increase the segmentation result. Automatic selection of the threshold(s) value becomes more difficult when the situation in image scene is even complicated, like overlaping objects or shadows. More precise preprocessing or previous knowledge about the objects is necessary [5,6].

### 2.3.1  Otsu

Otsu gray level thresholding is a nonparametric method of automatic threshold selection for picture segmentation from intensity histogram $H(p)$. Firstly the histogram functions is normalized:

$$o_p = \frac{H(p)}{N}$$

where $N$ is the total number of pixels in image.

For separating histogram into two classes, the probabilities of of class occurrence and the class mean are computed:

$$\omega_1 = \sum_{p=1}^{k} o_p, \quad \omega_2 = \sum_{p=k+1}^{L} o_p$$

$$\mu_1 = \frac{\sum_{p=1}^{k} p*o_p}{\omega_1}, \quad \mu_2 = \frac{\sum_{p=k+1}^{L} p*o_p}{\omega_2}$$

total mean level of image:

$$\mu_T = \sum_{p=1}^{L} p*o_p$$

and the between class variance:

$$\sigma_B^2 = \omega_1*(\mu_1-\mu_t)^2 + \omega_2*(\mu_2-\mu_t)^2$$

The optimal threshold $k^*$ maximazes $\sigma_B^2$ [8].

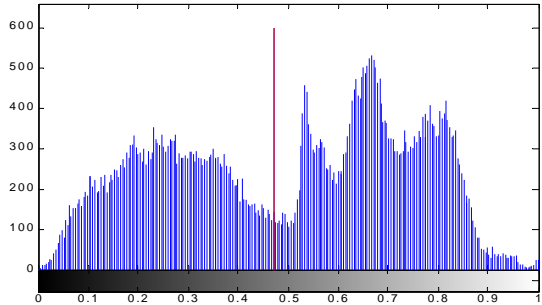

Figure 6: *Histogram function of Y1 with threshold from Otsu segmentation(red)*

## 2.4    Morphological operations

Forcing of objects structure based on nonlinear operations with *structural element*, the set of points, smaller then proceeded image. Structural element is moving across the image and the point of the image corresponding to coordinates of the structural element is changed in according to the relation betwen structral element and the image [5,6].

Dilation is the Minkowski addition of two sets $X$ and $SE$ and causes objects to grow in size and fill small holes inside.

$$\delta_{SE}(X) = X \oplus SE = \cup_{se \in SE}(X, se)$$

Erosion is a dual transformation to dilation, but not inverse function. It is Minkowski substraction and causes objects to shrink in size.

$$\epsilon_{SE}(X) = X \ominus SE = \cap_{se \in SE}(X, se)$$

Opening is an erosion followed by dilation and removes small objects.

$$(X \ominus SE) \oplus SE$$

Closing is a dilation followed by erosion and removes small holes.

$$(X \oplus SE) \ominus SE$$

## 2.4.1 Beucher gradient

Gradient operators are used in segmentation because they enhance high frequency intensity events in images. These events may be caused by edges of objects, but generally are caused by noise. The idea of real edges is somewhat fuzzy. Image objects are not necessary at real object boundaries. A cannonical definition of an edge is still to be found, if it exists.

Discrete case of Beucher gradient is defined as arithmetic difference between the dilation and the erosion of the image with the structural element B:

$$g_{SE} = \delta_{SE}(X) - \epsilon_{SE}(X)$$

Beucher gradient represents the maximum variation of the gray level intensity within structural element neighbourhood [9].

## 2.4.2 Grayscale reconstructin

Reconstruction extracts the connected components of image $X$ (mask) which are marked by image $M$ [10]. Both images are grayscale and of the same size. Value of each pixel from marker have to be smaller or equall to the pixel in mask on the same coordinates:

$$m(i, j) \leqslant x(i, j)$$

then the grayscale reconstruction can be defined as a geodesic dilation:

$$\rho_X^{(n)}(M) = min(\delta_X^{(n)}(M), X)$$

where $n$ is $n\text{-}th$ iteration, $\delta_X^{(n)}$ is dilation

$$\delta_X^{(n)}(M) = \rho_X^{(n-1)} \oplus SE$$

*SE* is morphological structural element and

$$\rho_X^{(0)}(M)=M$$

Stop condition for increasing *n* is when

$$\rho_X^{(n)}(M)=\rho_X^{(n-1)}(M) \ .$$

### 2.4.3 Watershed

Non-parametric detection method for closed contours. Grayscale image is considered as a topographic surface. Imagine a drop of watter on this topografic surgace. The watter streams down, reaches a minimum of height and stops there. Therefore the catchment basins and its watershed lines can be defined by means of a flooding process starting from thr minima. Let $p(i,j)$ be the value of a pixel on coordinates $(i,j)$. Then $M(p)$ is the set of the values in the neighborhood of $(i,j)$, that are stricly higher than $p(i,j)$. The point $(i,j)$) is constructible if and only if the nearest common ancestor of all the components in *M(p)*, has a value equal to or higher than $p+1$. The constructible point $(i,j)$ is iteratively selected, raised and updated until stability [11,12,13] .

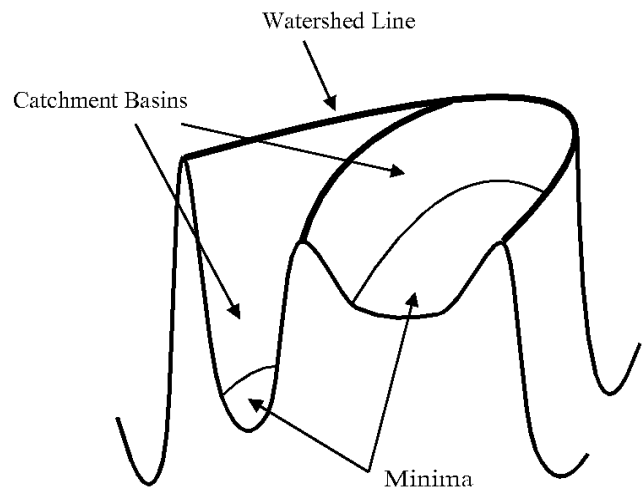

Figure 7.: *Flooding of a watershed, source http://cmm.ensmp.fr/*

## 2.5    Boundary curvature ratio

Merging criterion to improve the oversegmentation of watershed algorithm attempting to utilize boundary information to more visually appropriate segmentation. For two adjacent reginons $R_u$ and $R_v$ are computed the curvature ratio $k_i$.

$$k_i = \frac{\Delta x_i \Delta^2 y_i - \Delta^2 x_i \Delta y_i}{(\Delta x_i^2 + \Delta y_i^2)^{1.5}}$$

where $\Delta x_i = \frac{x_{i+1} - x_{i-1}}{2}$, $\Delta y_i = \frac{y_{i+1} - y_{i-1}}{2}$

$\Delta^2 x_i = \frac{\Delta x_{i+1} - \Delta x_{i-1}}{2}$, $\Delta^2 y_i = \frac{\Delta y_{i+1} - \Delta y_{i-1}}{2}$, and

$(x_i, y_i)$ is the coordinate position of $i^{-th}$ pixel in the contour.

Then, BCR of regions $R_u$ and $R_v$ is defined as:

$$BCR = \frac{\frac{1}{N_{u+v}} \sum_{i=1}^{N_{u+v}} k_i}{\frac{1}{N_u + N_v} \sum_{i=1}^{N_u} k_i + \sum_{i=1}^{N_v} k_i}$$

$N$ is the number of pixel in region.

If $BCR_{u,v} \geq 1$, the need to merge $R_u$ and $R_v$ is considered low and the cost to merge them is large. On the other hand, if $BCR_{u,v} \leq 1$, the shape integrity of merging them together is better then keep them separately and the cost to merge them is small [14].

# 3.    Object to background segmentation

Correct object found depends on plenty of factors, like kind of illumination, shadows, level of presented noise, proper focusing, overlapping to each other or objects dissimilarity to background. Founding process usually starts from simple techniques to more complicated algorithms, until the results are accetable.

For finding HeLa cells in the images from phase contrast microscope can be used the Otsu segmentation method. Because the original images are in RGB colour space representation, may be chosen one of grayscale transformations or splitting the colour channels into three different grayscale images – intensity of red, intensity of green and intensity of blue.

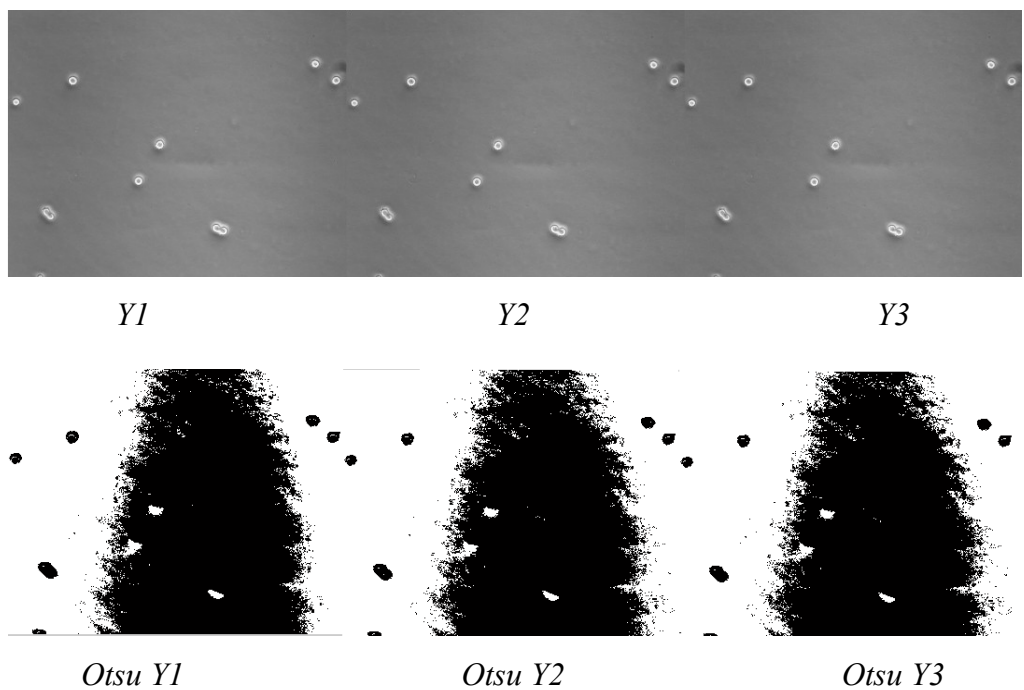

*Y1*                                    *Y2*                                    *Y3*

*Otsu Y1*                          *Otsu Y2*                          *Otsu Y3*

Figure 8.: Grayscale Otsu

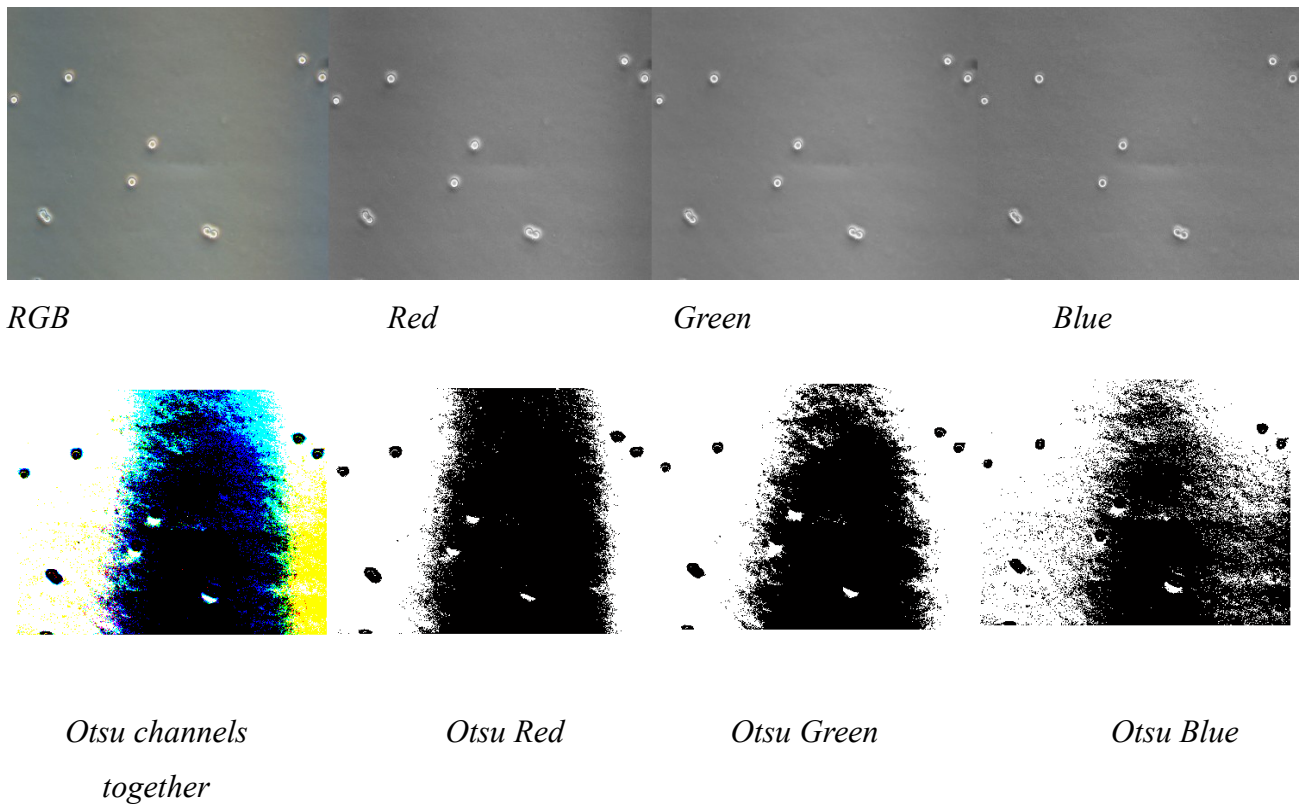

RGB                    Red                    Green                    Blue

Otsu channels            Otsu Red               Otsu Green               Otsu Blue
together

Figure 9.: *Splitting RGB and Otsu*

As is saw from Figure 8. and 9., direct using of Otsu did not tend to clear and present cells. There will be necessary to remove the effect of illumination and asperities on NUNC background.

## 3.1    Order statistic filtering

Order Statistic Filters are filters which rank neighboring pixels in an attempt to remove low frequency. The output pixels are computed by selecting a neighborhood of pixels around the input point. Then the pixels in the window are ranked according to their intensity. For minimum filter the minimal valued intensity is then assigned as the output value. For maximum filter the output value is the maximal valued intensity and for median filter it is the middle valued intensity. Large constant regions stay preserved and a thin one pixel line would be removed.

The order statistic filter is non-linear in its operation and effects. Higher statistics are used, the image will get brighter, while if lower statistics are used then the image gets darker.

Firstly for using the Filter on an input image, is important to set the right size of the neighborhood, the filtering window.

### 3.1.1  Appropriate range estimation

Any set of point can be divided into statisticaly appropriate number of equidistant intervals using one of these three  equations:

i)      $k = \sqrt{n}$

ii)     $k \leqslant 5 * \log_{10}(n)$

iii)    *The Sturges rule:* $k = 1 + 3.3 * \log_{10}(n)$

where *n* is number of points in set and *k* is the count of equidistant intervals [15].

Size of HeLa cell images is  $576 * 720 = 414720$ pixels per one image. It gives these possible numbers of intervals:

i)      $\sqrt{414720} \approx 645$

ii)     $5 * \log_{10}(414720) \approx 28$

ii)     $1 + 3.3 * \log_{10}(414720) \approx 20$

Because HeLa cells are quite small in comparison to the whole image, possibility i) was chosen. In that case, the number is also equal to the number of pixels in the window. Let choose the shape of it. As the basic shape is considered the square, but again according to the objects looked for, the circle shape of window could be better.

$$r = \sqrt{\frac{645}{\pi}} \approx 14$$

The filter window was determined as circle with radius *14* pixels.

### 3.1.2  Image border problem

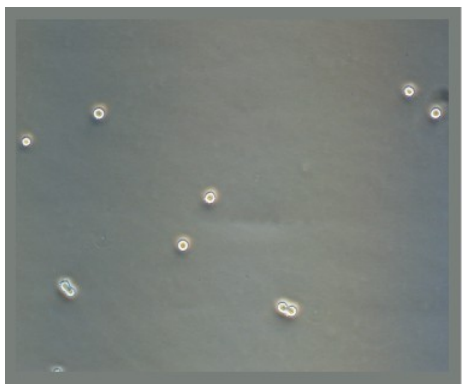

Figure 10: *Expanded Image borders*

On the all borders of the image only the half circle neighborhood is presented, even only quarter on the corners. As an extrapolation of the missing neighborhood, the size of whole image is on the borders expanded by area of thickness equal to the window radius. Value of extended area is computed as mean value of whole image [16].

### 3.1.3 Minimum Filter

Order statistical filtering with minimal value intensity parameter was separately applied on all channels of image with expanded borders. Then the borders were again removed.

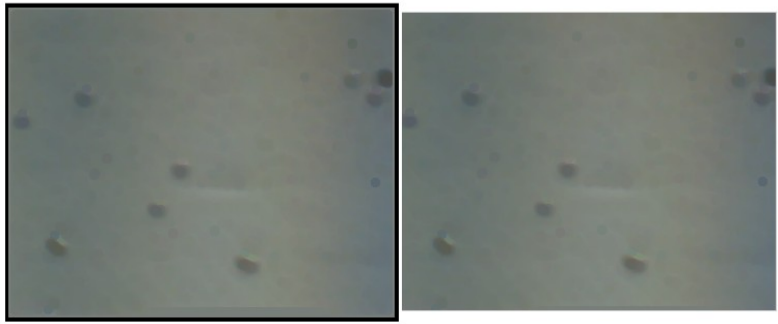

Figure 11.: *Minimal Filtered Image with and without extended borders*

Small asperities disappeared after minimal filtration, but this kind of filtering did not correct the influance of illumination. For removing it, more operations with the minimal filtered image should be done. Rate or difference between original and filtered image [7].

$$rate = \frac{filtered}{original} \qquad difference = original - filtered$$

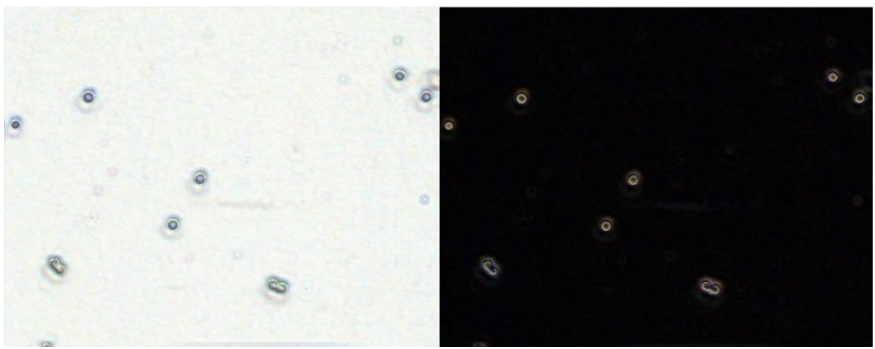

Figure 12.: *Rate(light) and difference(dark) between original and minimal filtered image*

### 3.1.4  Maximum filter

The same procedures with borders and filter window like in minimum filter were done for minimal value intensity. In this case in equations for rate and difference the *original* and *filtered* switched their places:

$$rate = \frac{original}{filtered} \qquad difference = filtered - original$$

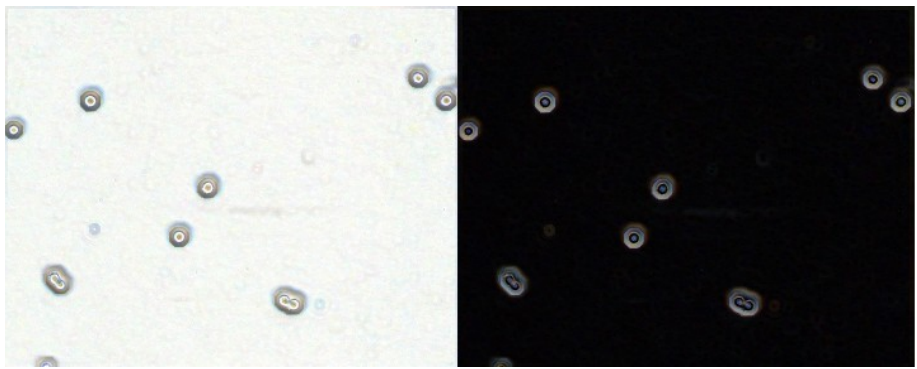

Figure 13.: *Rate(light) and difference(dark) between original and maximal filtered image*

### 3.1.5  Median Filter

Also, for median value intensity:

$$rate = \frac{original}{filtered} \qquad difference = original - filtered$$

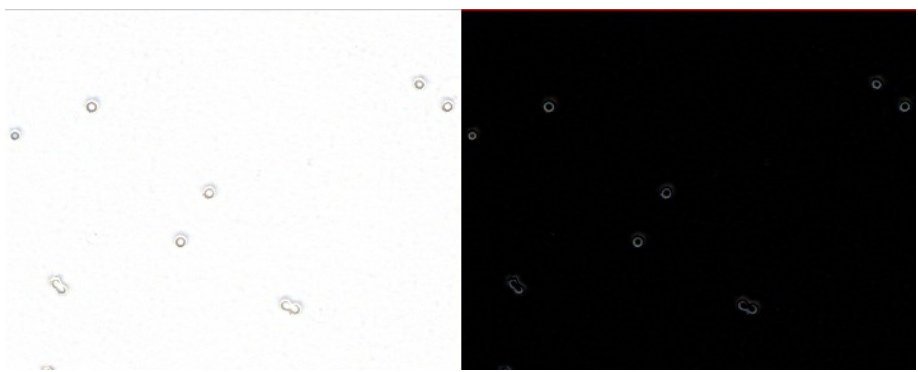

Figure 14.: *Rate(light) and difference(dark) between original and median filtered image*

### 3.1.6 Otsu after preprocessing

Now, when the filter functions were runing, the Otsu segmentation method for each channel should be tested again for all solutions from filtering, on both, rate and difference images, to choose which one is better.

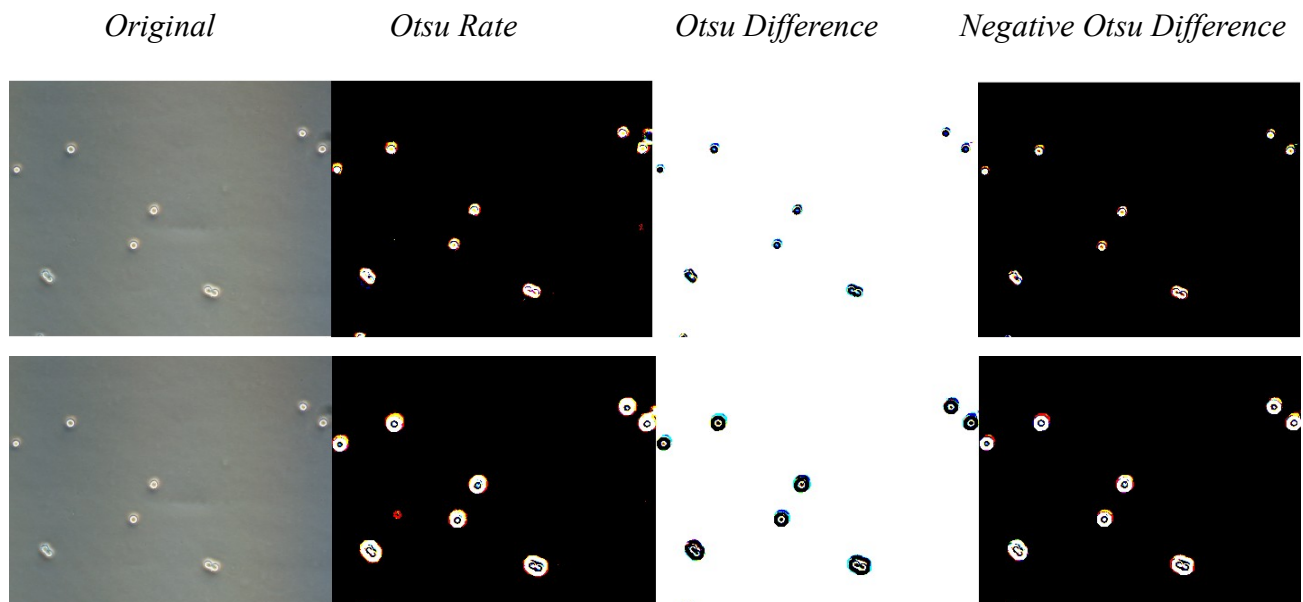

Figure 15.: *Otsu of Young HeLa cells, 1ˢᵗ minimal filter, 2ⁿᵈ* maximal filter

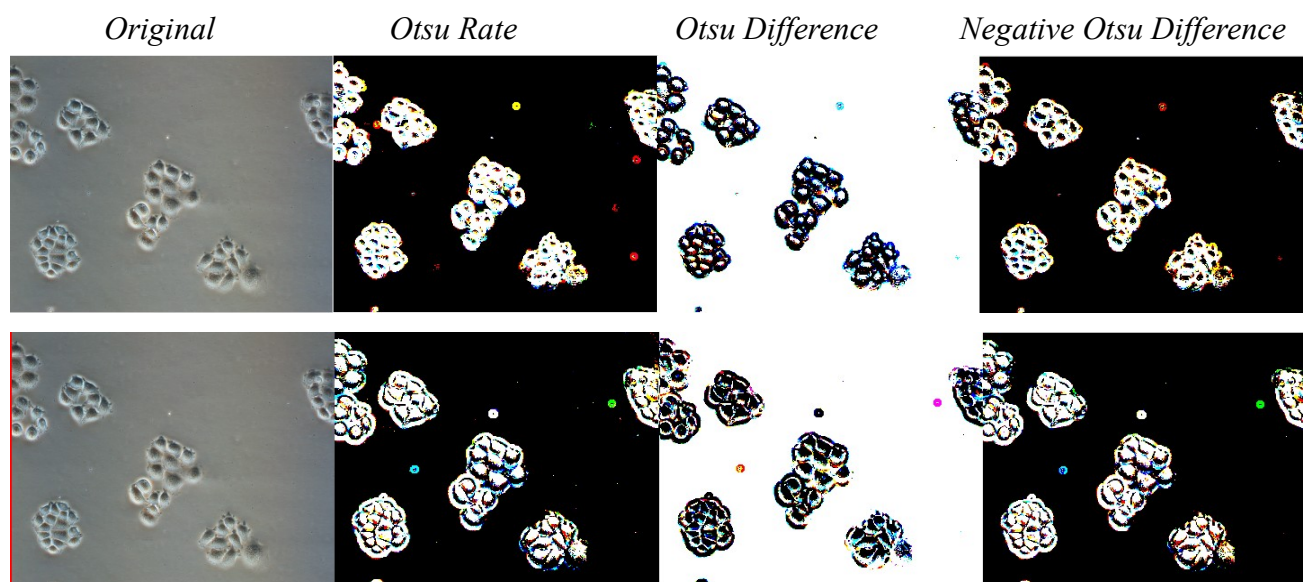

Figure 16.: *Otsu of Adult HeLa cells, 1ˢᵗ minimal filter, 2ⁿᵈ maximal filter*

*Original*                   *Otsu Rate*                   *Otsu Difference*

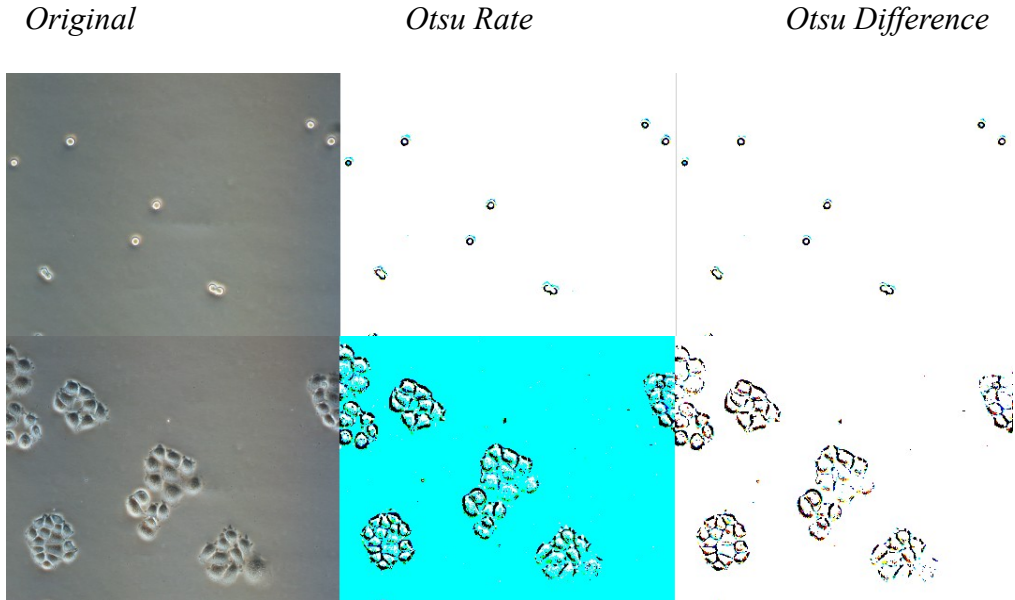

Figure 17.: *Otsu of median filtering, 1ˢᵗ Young HeLa, 2ⁿᵈ* Adult HeLa

Comparison of the results bringing the informations about usability of each filter method. Maximum filter is not so helpful because of emphasizing mainly the halo efects around the cells. Then this one was discard from another computation. Also in minimum filter a bit of halos remain. On the other hand, median filter emphasize only the borders of the cells.

Not decided yet if the rate could be better then difference, or the opposite statement.

## 3.2    Details emphasizing

After filtration, but before Otsu, still some improovement was missing. Three ways how to do it were tested. Let $p_{i,j,k}$ be intensity value of filtered pixel on coordinates *i, j* and channel *k*. Then $f_x(i,j,k)$ will be the new intensity value of the point $p_{i,j,k}$ :

i)      *prod:*     $f_a(i,j) = \prod_k (p_{i,j,k})$

ii)     *square:*  $f_b(i,j,k) = p_{i,j,k}^2$

iii)    *log:*     $f_c(i,j,k) = \sqrt{\mid \log_{10}(p_{i,j,k} + \frac{1}{512}) \mid}$

In case *prod* all channels are multiplicated together into grayscale image, where the dark pixels become more darker, and the light become more lighter.

Case *square* simply increase the contrast in image. The most complicated is case *log*. Because intensity values are between *0* and *1* , applying logarithm function will rapidly increase the differences between values. The domain range of logarithm function for values from interval *<0,1>* is *(-∞,0>*. For protection against negative infinity, the values were shifted up by small value 1/512. Absolut value operator was then used to convert negative values from logarithm function into positive. And finaly the radical was found, again to change the contrast.

### 3.2.1 Otsu after emphasizing

Otsu segmentation was repeated on filtered and emphasized imagas to see what changed.

*Emphasized prod*                     *Emphasized square*                    *Emphasize log*

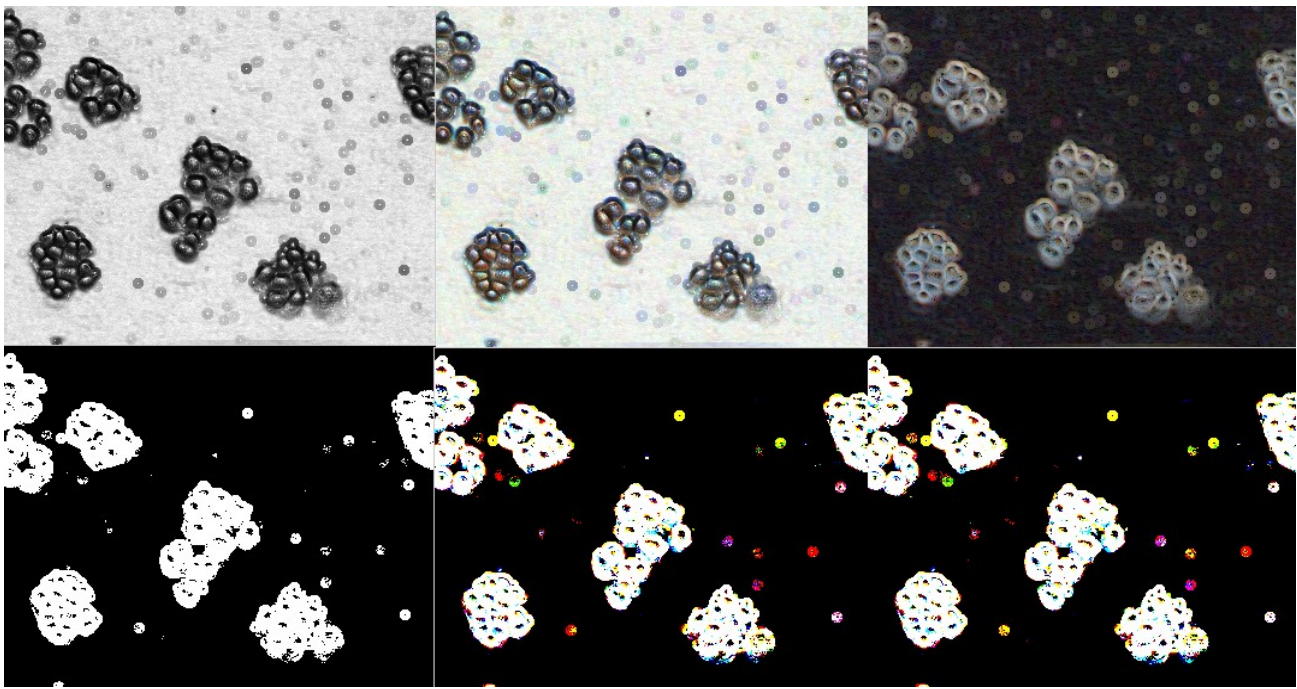

Figure 18.: Rate emphasized minimal filter and its Otsu

Is pretty clear that difference is not the proper comparison of original and filtered image, because of lot of disappered cells in several emphasizing and the highest occuring of fake segmented objects. In mimimal filtering the objects passing through the otsu segmentation are the borders of the HeLa cells, some borders between the cells and some borders inside the cells. The difference between kinds of emphasizing are only in the thickness of this borders. A little bit halo still rest.

*Emphasized prod*                  *Emphasized square*              *Emphasized log*

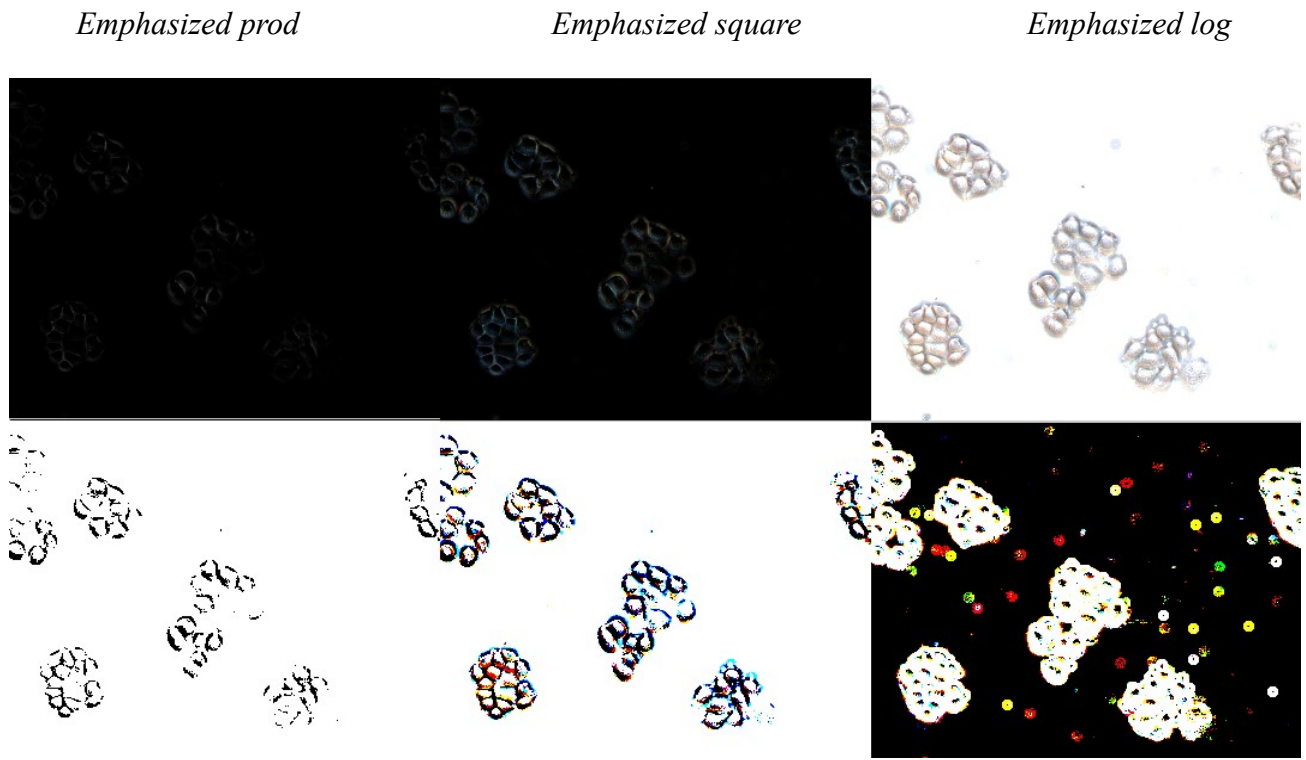

Figure 19.: *Difference emphasized minimal filter and its Otsu*

*Emphasized prod*                  *Emphasized square*              *Emphasized log*

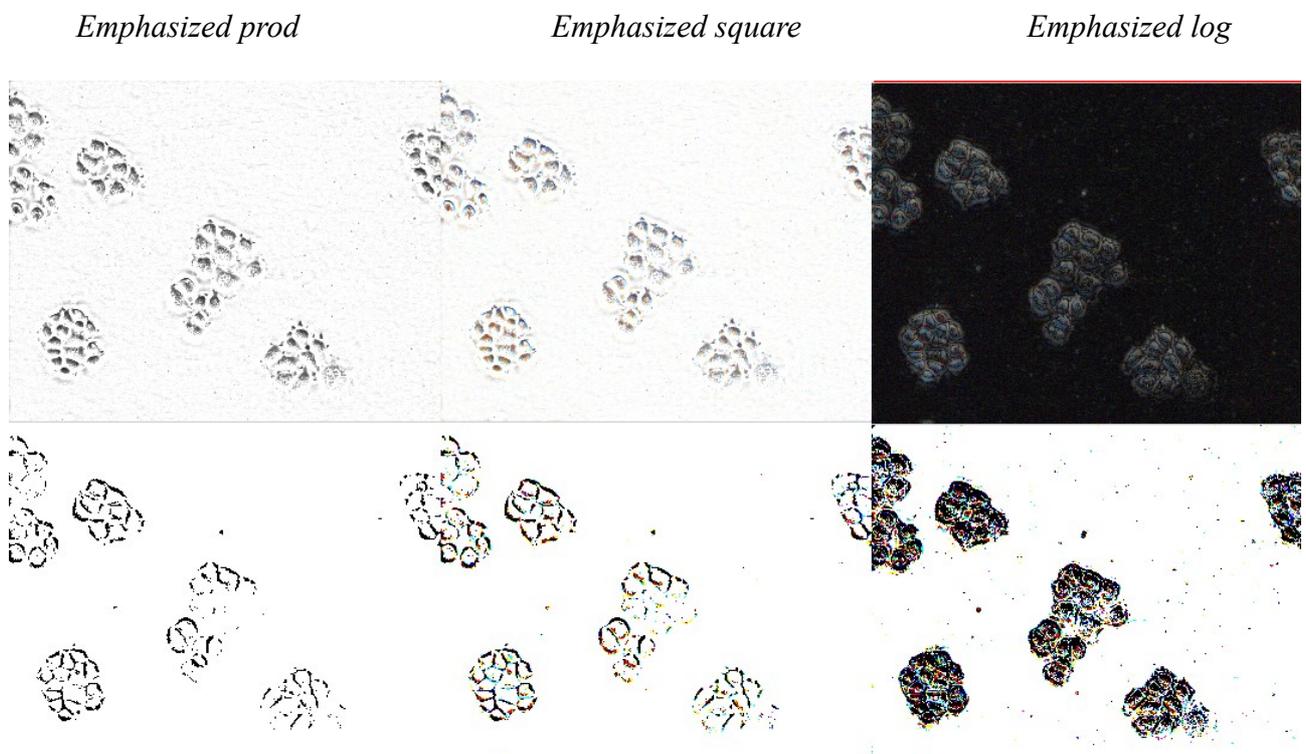

Figure 20.: Rate emphasized median filter and its Otsu

*Emphasized prod*          *Emphasized square*          *Emphasized log*

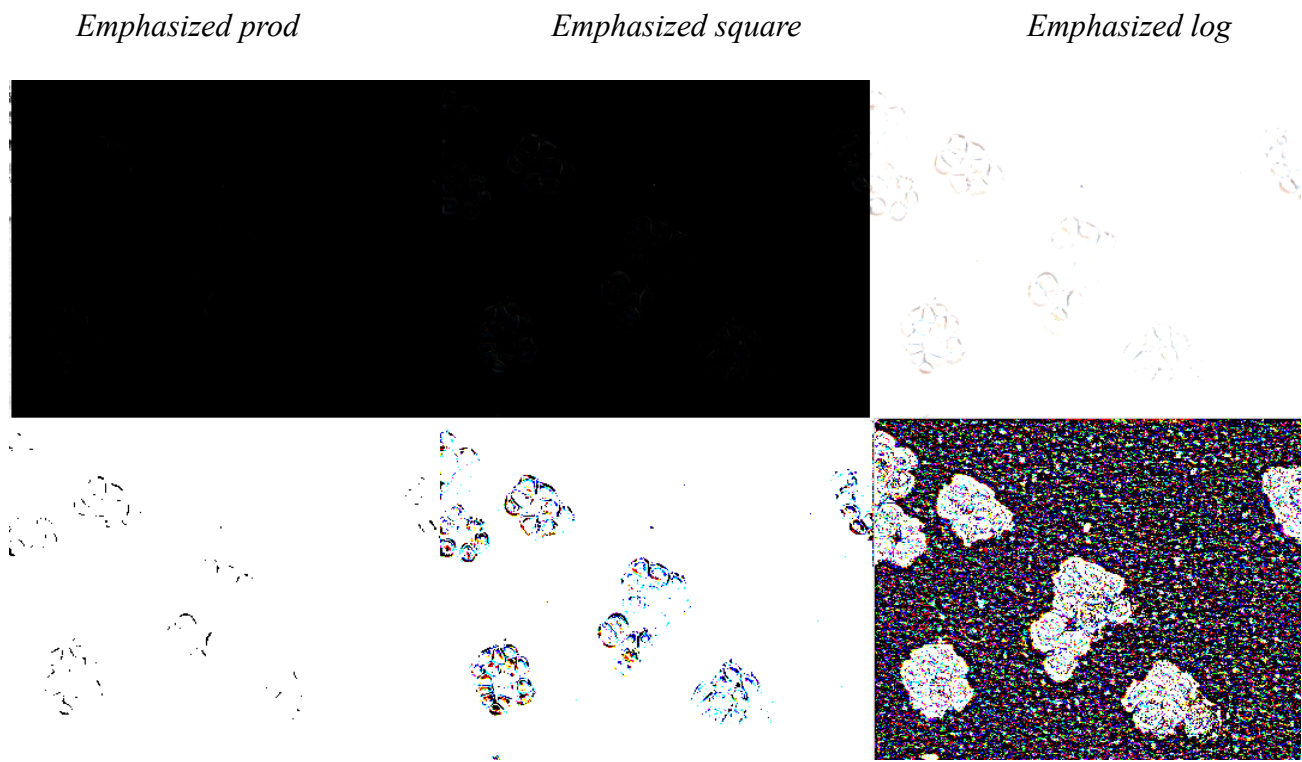

Figure 21.: *Difference emphasized median filter and its Otsu*

Generally, the logarithm emphasizing „survived" all types of filtration and comparison of original and filtrated images (rate or difference), so that probably makes it the most powerfull one of them. Moreover it is presented without halo in median filtering.

## 3.3   Smallest objects clearance

To remove objects that passed throught the Otsu segmentation but are not HeLa cells, some morphological operations could be used.

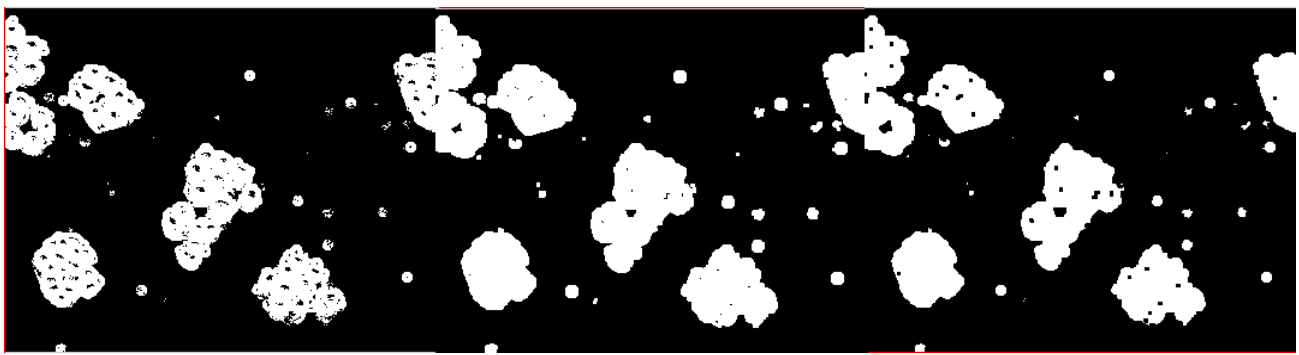

Figure 22.: *Otsu segmentation, its dilation and erosion*

The contour of the objects and holes were done by Beucher gradient [9] and for all arised lines computed their length. Simply lines shorter then the average length were deleted.

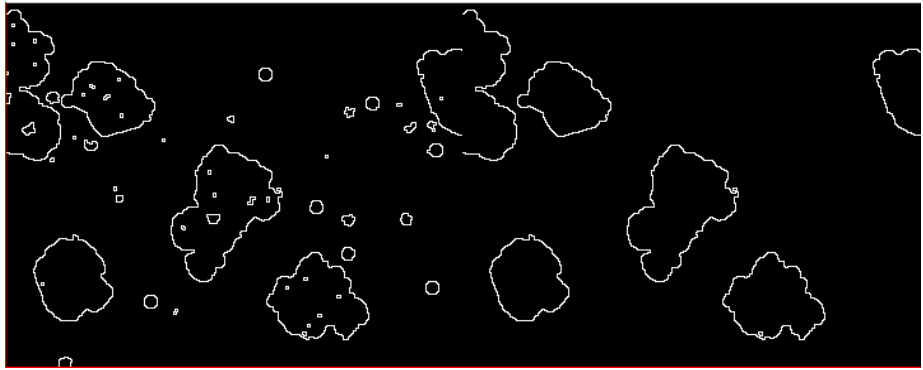

Figure 23.: *Beucher and Beucher without short lines*

From dilated image were counted all objects and computed they area. Similary like in the Beucher case, the objects with area less then average area, were deleted.

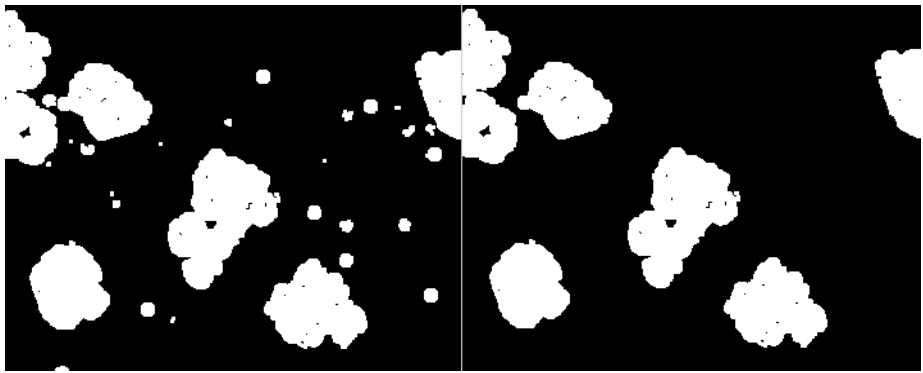

Figure 24.: *Dilation and dilation without small areas*

But it is still dilated image, that mean all objects are little bit larger then in the started Otsu segmentation. Using dual erosion could created bigger holes inside the objects. Better solution, how to shrink the objects is to substract the remaining Beucher gradient lines from remining dilation.

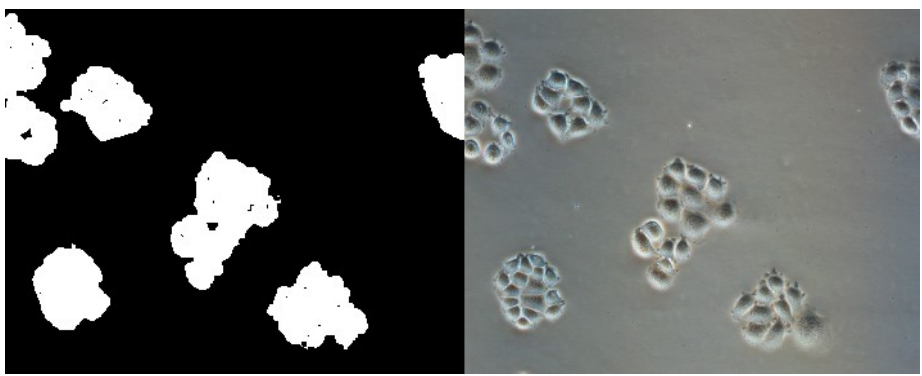

Figure 25.: Cleared objects and original image

## 3.4    Binary image

Result form *prod* emphasizing are in binary image *(*value equal *0* for NUNC background and *1* for cells*)* and could be directly used as a mask. But in *square* and *log* emphasizing, the results are three binary images – one for each channel.  Let  $p_{i,j,k}$ be the point on coordinates *i,j* and channel *k*. Then the point in final binary image,  is $bw_{i,j}$. There are three ways how to make a binary image from three binary channels:

i)      *strict:*            $bw_{i,j} = \prod_k p_{i,j,k}$

ii)     *patient:*          $bw_{i,j} = 1 \quad if\,(\sum_k p_{i,j,k}) > 0$
                              $= 0 \quad otherwise$

iii)    *halfway:*         $bw_{i,j} = 1 \quad if\,(\sum_k p_{i,j,k}) > 1$
                              $= 0 \quad otherwise$

In *strict* case, the point that is considered as object in all channels together become object point in final binary image. Some points may be lost, because they have not to be presented in all channels congruously. On the other hand, the *patient* case, allows to all possible points from all channels, to be correctly segmented object points. That can make a lot of fake points. In equilibrium between first two cases is the *halway* one, where only points that are in at least two channels, remain in the final binary image. There is lower level of lost points than in *strict* case, and also lower level of fake points than in *patient* case.

| *3-channels* | *strict* | *patient* | *halfway* |
| --- | --- | --- | --- |

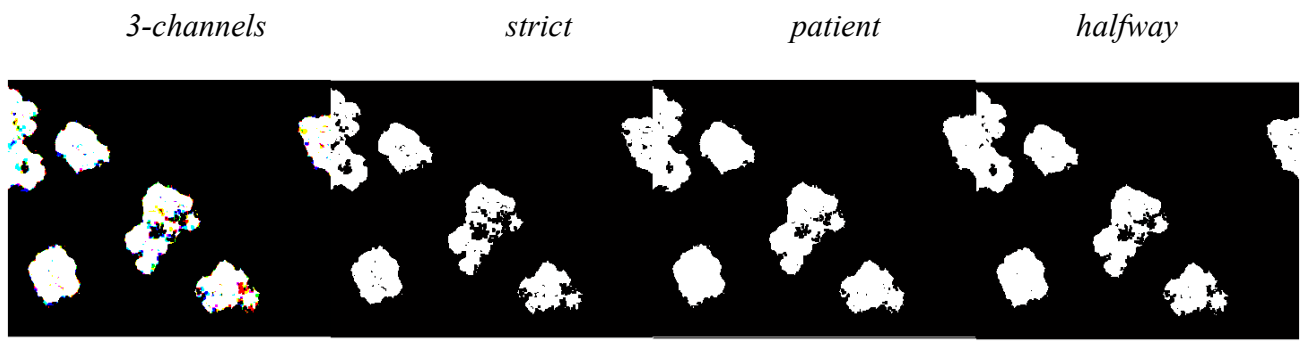

Figure 26.: *Cleared Otsu segmentation of  log emphasized rate of median filtered image*

## 3.5    Cells to background recap

Various precesses how to suceed in segmentation were tested and reviewed. Shown how the sequence of the algorithms were found. Several steps, constisted of described methods and some their improovements, were defined:

i)       rate between original image and minimum or median filtration, with estimation of proper structural element.

ii)      details emphasizing by using log method

iii)     statisticaly based object clearance after otsu segmentation on preprocessed image

iv)     halfway binarizing of three channel binary segmented images

After median filtration with log emphasizing and before otsu segmentation have to be done negative transformation to switch the *0* and *1* values into binary opposite.

Presented steps were running on randomly chosen images to prove their abilities.

<div align="center">
<i>Original</i>            <i>mimimum</i>            <i>median</i>
</div>

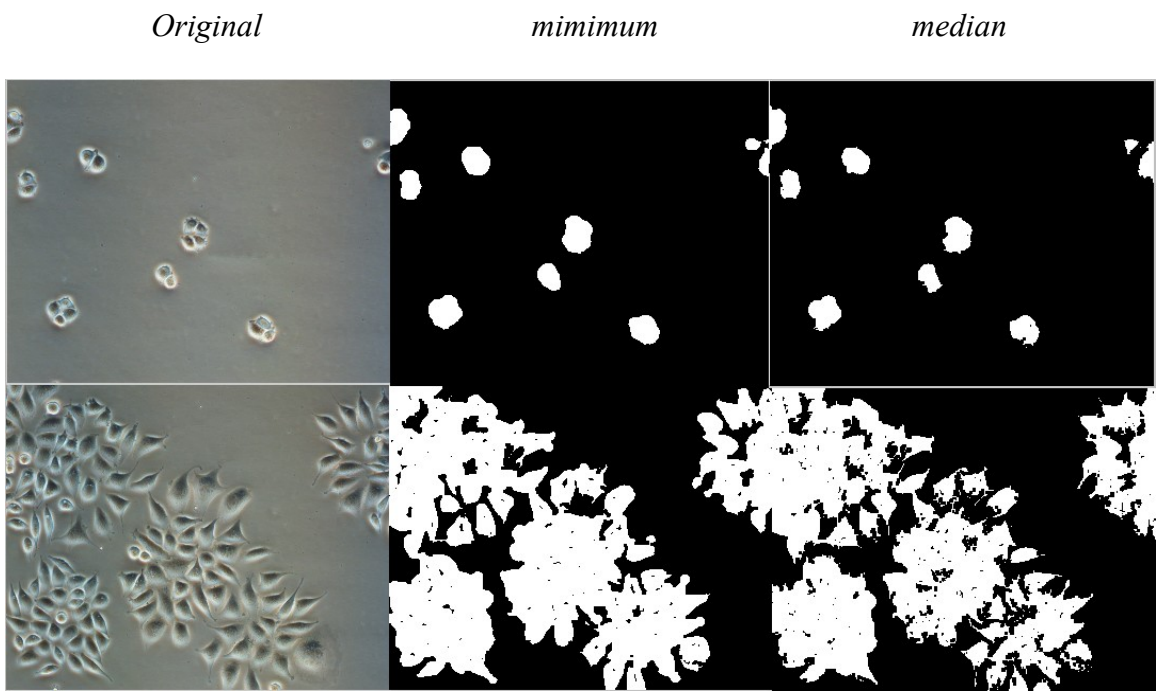

Figure 27.: *Young and Adult HeLa cells to background segmentation.*

Both filters, minimum and media, give similar results, but in minimum filter still remain the halo efect. So the more accurate is to using median filtration rate with logarithmic emphasizing before otsu segmentation folowed by clearence and binarization.

# 4  Independent cells location

Result from object to background segmentation was a mask, that can be multiply by original image to get only the cells on dark background. Looking for individual cells and separating them to each other would by done on this masked image.

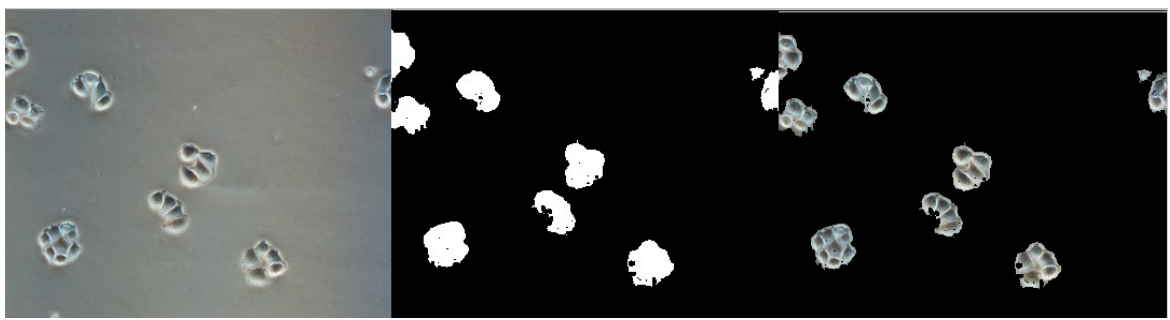

Figure 28.: *Application of mask.*

One of the simplest contours detection is the Beucher gradient. When it is applied on masked image, the outer boundary of cells clusters were pretty clear. But only some inner boundary little bit emphasized and some non-boundary pixels joined the show.

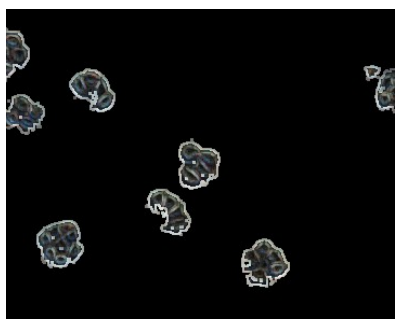

Figure 29.: *Beucher gradient of masked image*

## 4.1  Watershed morphology

More powerful countour detection method is to use the watershed, which is independent on shape and countour, efficient and accurate. Grayscale image is, according to the method theory, divided into two parts – the catchment basins and the watersheds.

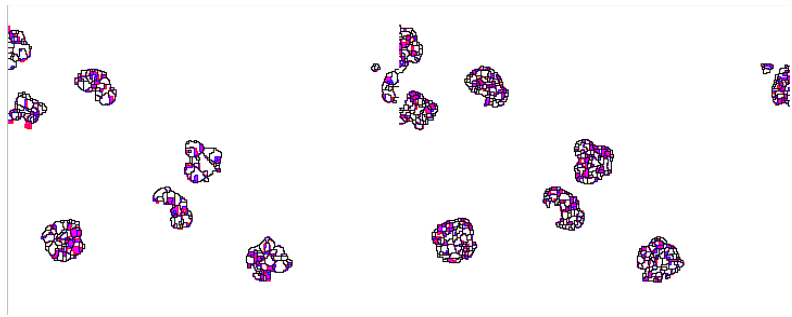

Figure 30.: *Watershed and watershed on Beucher gradient,*
*white for catchment basins, black for watershed lines,*
*other for basins in some channel and line in another*

The main problem of watershed method is the oversegmentation (see Figure 30.). In real images with structured objects, it is able to find a lot of catchment basins and watershed lines, that are not actually associated with meaningful regions. Even watershed on Beucher gradient is oversegmented much more.

## 4.2    BCR for adjacent segments

To reduce number of segments after watershed segmentation, the boundary curvature ratio was presented as improovement method. It compute the cost in shape integity of two adjacent regions. Firstly, from is necessary to find if two regionts $R_u$ and $R_v$, resulted from watershed segmentation are adjacent or not.

$$\delta R_u = R_u \oplus SE$$

$$\delta R_v = R_v \oplus SE$$

$$\cap (\delta R_u, \delta R_v) > 0$$

if the intersection of segments dilation is not an empty set, then they are adjacent and their boundary curvature ratio can be computed. As the pixels of boundary are considered pixels from Boucher gradient of watershed segmentation (do not mismatch with watershed of Boucher gradient, it differ in order of operations).

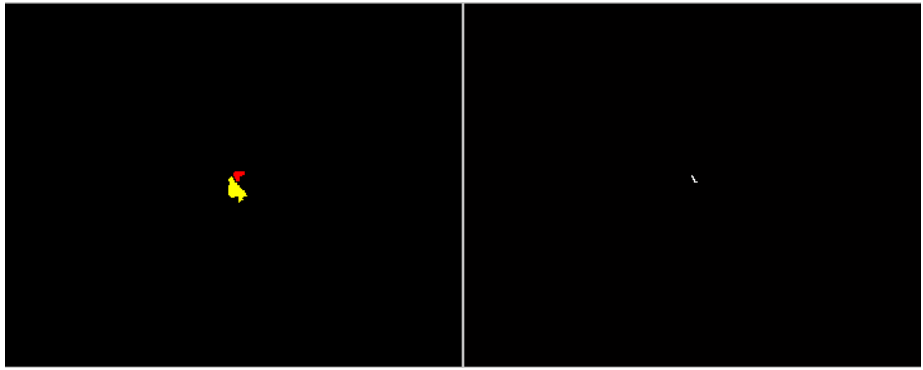

Figure 31.: *Region 191 (yellow) with region 193 (red), and intersection (white) of their dilations*

After computing all boundary curvature ratios, the regions that belongs together according to their BCR were merged and colorized (see Figure 32.). Unfortunately also the regions, that not belongs together were merged and that show, the BCR technique is not adequate enough for small regions of same types but different shapes.

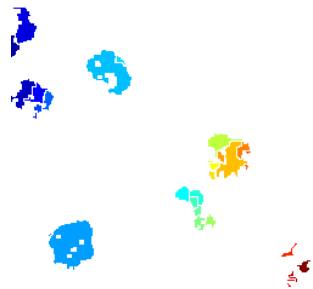

Figure 32.: *Region merging after BCR*

## 4.3    Opening-by-reconstruction and closing-by-reconstruction

Opening-by-reconstruction is erosion followed by morphological reconstrutin, closing-by-reconstruction is dual operation to opening-by-reconstruction and it is dilation followed by morphological reconstruction. Both techniques are used to clean up the image. They could remove small blemishes without affecting the overall shapes of the objects [10].

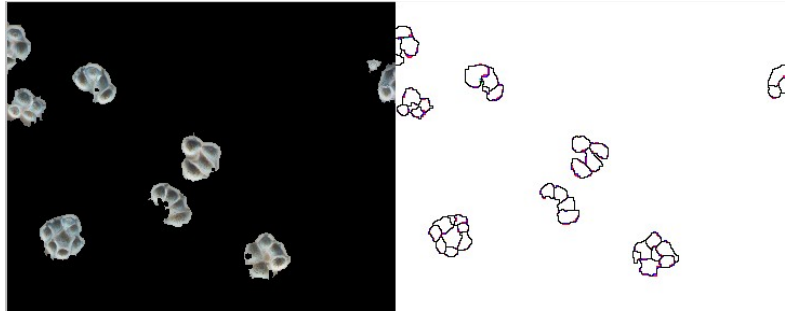

Figure 31.: *Opening-by-reconstruction followed by closing-by-reconstruction and its watershed*

## 4.4    Independent cells location recap

Watershed segmentation is powerfull method for separating image into regions. The oversegmentation can be reduced by applying opening-by-reconstruction and closing-by-reconstruction before using watershed.

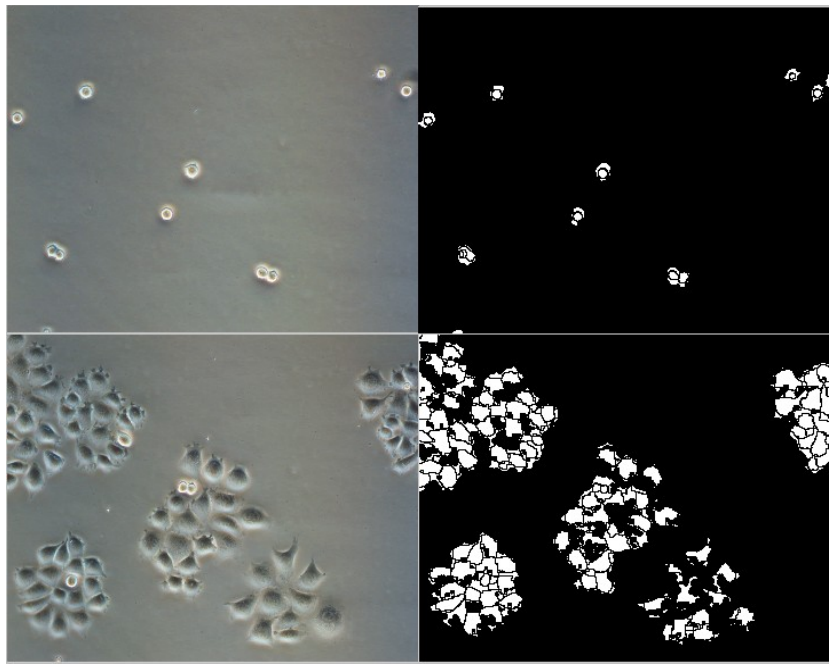

Figure 32.: *Final segmentation of  Young (up) and Adult (down) HeLa cells*

# 5    Output parameters

The most important parameter for biological point of view is the count of spheric and nonspherc cells in the image. To count the number of objects is simple, but for counting different types is necessary to separate them before. Some kind of sphericity have to be computed as a treshold for this separation. How to obtain the sphericity?

The perimeter of object can be easily reached, for example from Beucher gradient. Also the area of objects as the number of pixels in object should be known. Now, from the sphere feature:

$$\text{radius from perimeter: } r_p = \frac{perimeter}{2 * \pi}$$

$$\text{radius from area: } r_a = \sqrt{\frac{area}{\pi}}$$

$$\text{for real sphere is true } \frac{r_p}{r_a} = 1$$

The one of the most important feature of sphere is that it has the largest area with given perimeter, or the smallest perimeter wirh given area. Then as the sphericity can be considered rate between $r_p$ and $r_a$. How close this rate is to *1*, is how close the shape is to sphere.

The treshold for sphere was set to 1.1.

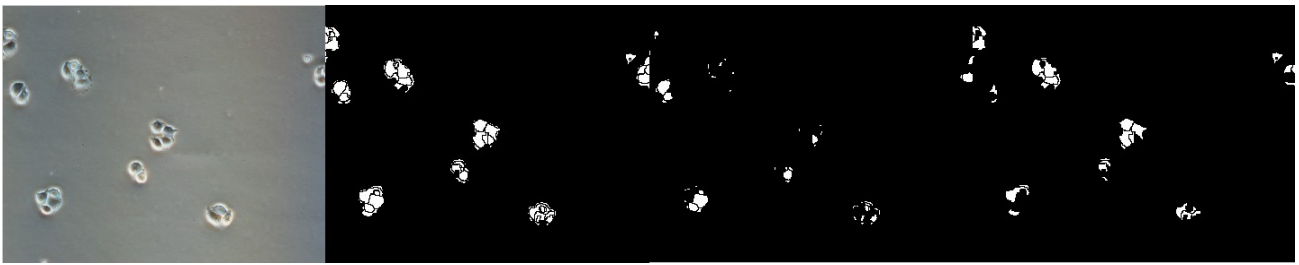

Figure 33.: *Original image, segmentation, spheric cells, nonspheric*

# 6    Conclusion

Several techniques for segmentation were tested. Cells to background segmentation were completly successful for both types of cells, young or adult. Otsu segmentation is very powerfull method for binary segmentation, but in this case needs polished preprocessing. Local median filtration, with described appropriate range estimation for structural element, and following logarithmic details emphasizing, was used to improove information differnciability in image, before appluing Otsu segmentation.

For independent cells loacation, watershed methods was helpfull but with all problems of this methods. Boundary curvature ratio can not solve the oversegmantation in this case, probably because of poor dissimilarity of independent cell. Applying opening-by-reconstruction and closing-by-reconstruction, again with appropriate range estimation for structural element, solve lot of oversegmentation, but still small amount of oversegmented or nonsegmented cells remain. Accurate segmentation technique is still something that is looking for.

The last part of given task, was to express the sphericity of cells, simple but proper algorithm was described and used, even against the small errors from watershed segmentation.

The algorithms are able to be used in praxis, only tresholding for possibility of cell size have to be done later, to remove noises that were recognized as cells, but are under their size. Comparing by average values or standard deviations of the values of pixel din not solve it.